\documentclass{article}

\usepackage{microtype}
\usepackage{graphicx}
\usepackage{subcaption}
\usepackage{booktabs} %

\usepackage{hyperref}

\usepackage[accepted]{icml2021}

\usepackage{cleveref}
\usepackage{amsfonts}

\graphicspath{{figs/}}

\icmltitlerunning{Deep Deterministic Uncertainty for Semantic Segmentation}

\begin{document}

\twocolumn[
\icmltitle{Deep Deterministic Uncertainty for Semantic Segmentation}

\icmlsetsymbol{equal}{*}

\begin{icmlauthorlist}
\icmlauthor{Jishnu Mukhoti}{oatml,tvg}
\icmlauthor{Joost van Amersfoort}{oatml}
\icmlauthor{Philip H.S. Torr}{tvg}
\icmlauthor{Yarin Gal}{oatml}
\end{icmlauthorlist}

\icmlaffiliation{oatml}{Oxford Applied \& Theoretical Machine Learning Group, Department of Computer Science, University of Oxford, Oxford, United Kingdom}
\icmlaffiliation{tvg}{Torr Vision Group, Department of Engineering Science, University of Oxford, Oxford, United Kingdom}
\icmlcorrespondingauthor{Jishnu Mukhoti}{jishnu.mukhoti@eng.ox.ac.uk}

\icmlkeywords{Machine Learning, ICML}

\vskip 0.3in
]

\printAffiliationsAndNotice{} %

\begin{abstract}
We extend Deep Deterministic Uncertainty (DDU) \citep{mukhoti2021deterministic}, a method for uncertainty estimation using feature space densities, to semantic segmentation.
DDU enables quantifying and disentangling epistemic and aleatoric uncertainty in a single forward pass through the model.
We study the similarity of feature representations of pixels at different locations for the same class and conclude that it is feasible to apply DDU location independently, which leads to a significant reduction in memory consumption compared to pixel dependent DDU.
Using the DeepLab-v3+ architecture on Pascal VOC 2012, we show that DDU improves upon MC Dropout and Deep Ensembles while being significantly faster to compute.
\end{abstract}

\vspace{-5mm}
\section{Introduction}
\label{sec:introduction}

With the increasing deployment of deep learning models in safety critical applications like autonomous driving \citep{huang2020autonomous} and medical diagnosis \citep{esteva2017dermatologist}, it is imperative for such models to be able to quantify their uncertainty reliably, in addition to making accurate predictions.
A significant amount of research has been conducted in this direction and several methods have been introduced in the context of classification \citep{gal2016dropout, deep_ensembles, blundell2015weight}.
These methods require several forward passes through the model rendering such methods practically infeasible for adoption in large scale applications like semantic segmentation \citep{long2015fully}, where dense pixel-wise predictions are necessary, often in real time.

Recently, several methods have been introduced to obtain uncertainty in a single forward pass \citep{van2020uncertainty, liu2020simple, mukhoti2021deterministic}.
In particular, DUQ \citep{van2020uncertainty} and SNGP \citep{liu2020simple} propose using feature extractors with certain inductive biases to impose a bi-Lipschitz constraint on the feature space. They then use a distance aware layer, either an RBF or a Gaussian Process trained end-to-end with the feature extractor. However, both these methods require extensive changes to the model architecture and training setup, with additional hyperparameters which need to be fine-tuned. DDU \citep{mukhoti2021deterministic} shows that using feature space density with proper inductive biases can capture uncertainty and avoids the problem of \emph{feature collapse} \citep{van2020uncertainty}. Due to feature collapse, Out-of-distribution (OoD) samples are often mapped to in-distribution regions in the feature space, making the model overconfident on such inputs.
Hence, in order to capture uncertainty through feature space density, one needs to use proper inductive biases on the model architecture.

There are two kinds of uncertainty which are important in deep learning literature: \emph{epistemic uncertainty}, which captures what the model does not know, is high for unseen or OoD inputs and can be reduced with more training data and, \emph{aleatoric uncertainty}, which captures ambiguity and observation noise in in-distribution samples \citep{kendall2017uncertainties}. In DDU, the epistemic uncertainty is quantified using a feature space density, while the entropy of the softmax distribution can be used to estimate aleatoric uncertainty.

\begin{figure*}[t]
    \centering
    \includegraphics[width=0.7\linewidth]{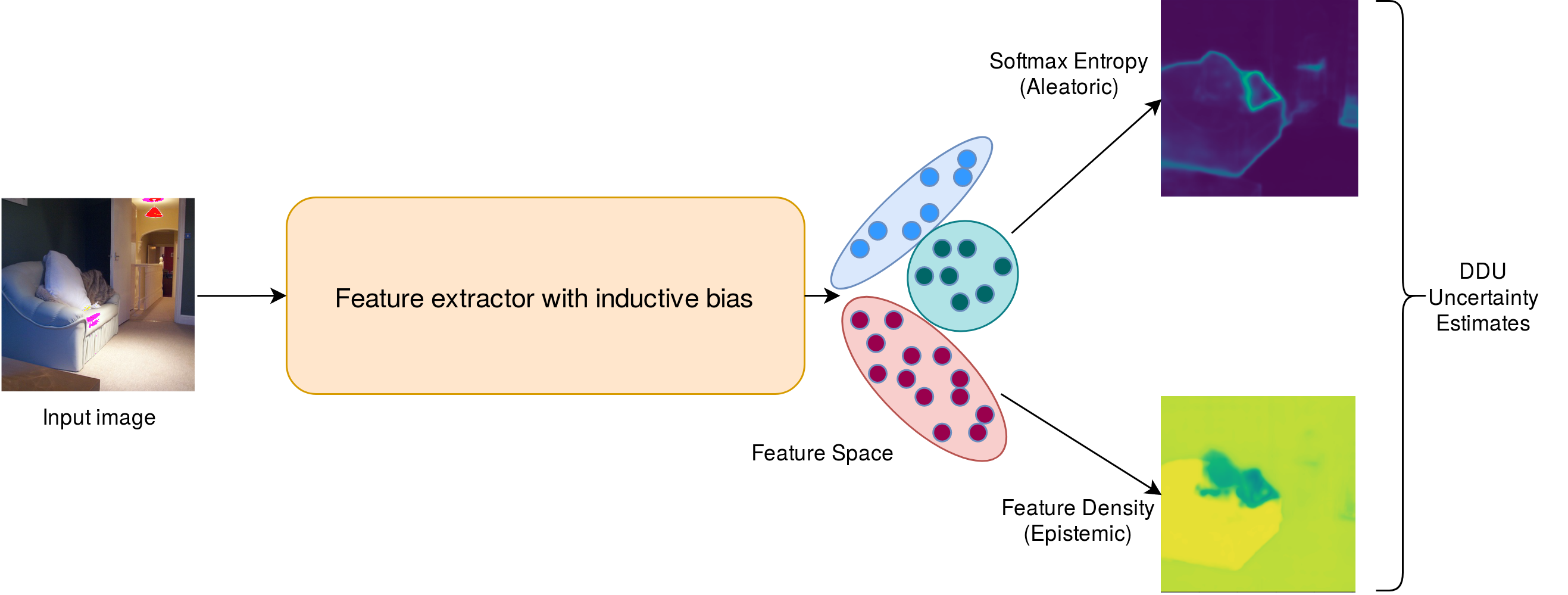}
    \caption{Applying DDU in the context of semantic segmentation}
    \label{fig:ddu_semsegs}
\end{figure*}

In this paper, we apply and extend DDU to the task of \emph{semantic segmentation} \citep{long2015fully}, where each pixel of a given input image is classified to produce an output which has the same spatial dimensions as the input. We choose semantic segmentation in particular as it forms an excellent example of an application with class imbalance and therefore, requires reliable epistemic uncertainty estimates. Furthermore, state-of-the-art models for semantic segmentation \citep{chen2017rethinking, zhao2017pyramid, wang2020deep} are large and conventional uncertainty quantification methods like MC Dropout \citep{gal2016dropout} and Deep Ensembles \citep{deep_ensembles} are often prohibitively expensive on such models \citep{kendall2015bayesian, mukhoti2018evaluating}.

The paper is organised as follows: in \Cref{sec:ddu_semseg}, we describe how DDU can be extended for semantic segmentation and in \Cref{sec:experiments}, we provide results using the well-known DeepLab-v3+ \citep{chen2017rethinking} architecture on the Pascal VOC 2012 \citep{everingham2010pascal} dataset to show that DDU outperforms other conventional methods (MC Dropout and Deep Ensembles) of uncertainty quantification in deep learning.

\section{DDU in Semantic Segmentation}
\label{sec:ddu_semseg}

In this section, we provide details on how DDU can be extended to obtain epistemic and aleatoric uncertainty estimates in semantic segmentation.

\textbf{A brief introduction to DDU:} As described in \citet{mukhoti2021deterministic} in the context of multiclass classification, after training a model with a bi-Lipschitz constraint, we can compute the feature space means and covariances per class using a single pass over all the training samples. The feature space means and covariance matrices can then be used to fit a Gaussian Discriminant Analysis (GDA) \citep{murphy2012machine}. Let $z$ be the feature representation for a given input $\mathbf{x}$, i.e., $z = f_\theta(\mathbf{x})$ where $\theta$ represents model parameters. Then the feature density $p(z)$ is computed by marginalizing the density over all classes as
\begin{equation}
\label{eq:density}
    p(z) = \sum_c p(z|c)p(c)
\end{equation} where $p(z|c)$ is obtained from the GDA and $p(c)$ can be computed directly from the training set. The feature density thus computed can be used to estimate model confidence (opposite of epistemic uncertainty). At the same time, for in-distribution samples, the entropy of the softmax distribution $\mathbb{H}[p(\mathbf{y|x, \omega})]$ can be used to capture aleatoric uncertainty.

\begin{figure}[t!]
    \centering
    \begin{subfigure}{0.3\linewidth}
        \includegraphics[width=\linewidth]{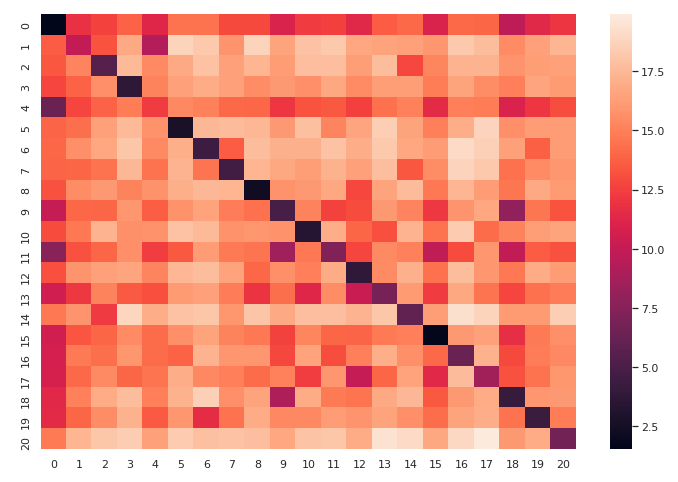}
        \label{fig:p1_matrix}
    \end{subfigure}
    \begin{subfigure}{0.3\linewidth}
        \centering
        \includegraphics[width=\linewidth]{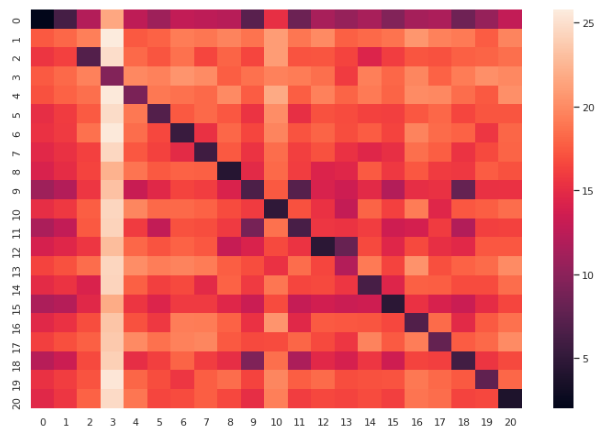}
        \label{fig:p2_matrix}
    \end{subfigure}
    \begin{subfigure}{0.3\linewidth}
        \centering
        \includegraphics[width=\linewidth]{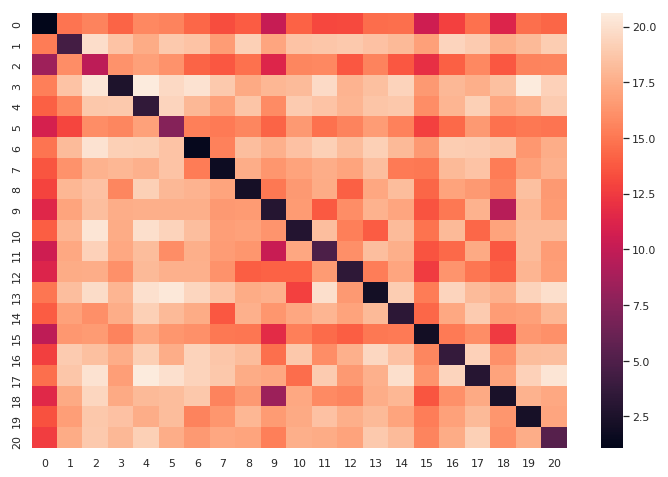}
        \label{fig:p3_matrix}
    \end{subfigure} 
    \vspace{-3mm}
    \caption{
    L2 distances between the feature space means of different classes for a pair of distant pixels on the Pascal VOC 2012 val set: (left) Pixels $(10, 255)$ and $(500, 255)$, (middle) Pixels $(234, 349)$ and $(36,22)$ and (right) Pixels $(300, 500)$ and $(400, 255)$.
    }
    \label{fig:l2_dist}
    \vspace{-5mm}
\end{figure}

\textbf{Pixel-independent class-wise means and covariances:} In semantic segmentation, each pixel has a prediction attached to it and a corresponding softmax distribution. A natural question to ask then is whether to compute means and covariance matrices per pixel in order to fit a GDA to the semantic segmentation model. Fortunately, we find that such is not the case and we can indeed compute means and covariance matrices independent of pixels just like in multi-class classification (thereby enforcing invariance). To see this, in \Cref{fig:l2_dist}, we plot the L2 distances between the feature space means of all pairs of classes in the Pascal VOC validation set for two ``distant'' pixels. We find that the means of the same class are much closer together as compared to other classes irrespective of where the pixels are located. This makes intuitive sense as the convolution kernel (a linear operator) which converts the feature space representations into logits is shared across the entire feature space representation.

\textbf{Computing feature density:} Following the rationale above, we fit a GDA assuming pixels to be independent samples. Hence, we obtain one mean and one covariance matrix per class (not per pixel) and can apply \Cref{eq:density} to obtain the feature density per pixel, given an input image. Separately, we can also obtain the per pixel softmax entropy from the model. Using these two we can disentangle aleatoric and epistemic uncertainty with a single deterministic model in semantic segmentation. We present a schematic diagram of this process in \Cref{fig:ddu_semsegs}.

\section{Experiments}
\label{sec:experiments}

In this section, we evaluate DDU on semantic segmentation using the well-known Pascal VOC \citep{everingham2010pascal} dataset and compare it with three other uncertainty baselines widely applied in practice: softmax entropy, MC Dropout \citep{gal2016dropout} and Deep Ensembles \citep{deep_ensembles}.

\textbf{Architecture \& Training setup:} For all the baselines, we use DeepLab-v3+ \citep{chen2017rethinking} using a ResNet-101 \citep{he2016deep} backbone as the architecture of choice. We train each model for 50 epochs on the Pascal VOC training set augmented using the Semantic Boundaries Dataset (SBD) \citep{BharathICCV2011} using SGD as the optimiser with a momentum of $0.9$ and a weight decay of $5e-4$. We set the initial learning rate to $0.007$ with a polynomial decay during the course of training. Finally, we trained using a batch size of $32$ parallelized over 4 GPUs.

\textbf{Baselines \& Uncertainty metrics}: As mentioned before, we compare DDU with 3 well-known baselines:

\begin{table}[t!]
    \centering
    \begin{tabular}{c|c|c}
        \toprule
        \textbf{Baseline} & \textbf{mIoU} & \textbf{Runtime (ms)} \\
        \midrule
        Softmax & $78.53$ & $275.48\pm1.91$ \\
        MC Dropout & $78.61$ & $1576.75\pm1.56$ \\
        Deep Ensemble & $78.47$ & $875.87\pm0.79$ \\
        \textbf{DDU} & $78.53$ & $\mathbf{263.83\pm2.79}$ \\
        \bottomrule
    \end{tabular}
    \caption{\emph{Pascal VOC val set mIoU and runtime in milliseconds of a single forward pass for different baselines averaged over 10 single forward passes.} Note that for each single forward pass in the MC Dropout baseline, we perform 5 stochastic forward passes.}
    \label{tab:runtime}
\end{table}

\begin{enumerate}
    \item \emph{\textbf{Softmax Entropy}}, one of the most commonly used metrics for uncertainty is the entropy $\mathbb{H}[p(\mathbf{y|x, \omega})]$ of the softmax distribution $p(\mathbf{y|x, \omega})$ \citep{hendrycks2016baseline}. This metric is often preferred due to its simplicity and lack of computational overhead. Softmax entropy is known to capture aleatoric uncertainty for in-distribution samples \citep{mukhoti2021deterministic}. However, it cannot capture epistemic uncertainty reliably (eg. for OoD inputs).
    
    \item \emph{\textbf{MC Dropout (MCDO)}} \citep{gal2016dropout} is a method which uses dropout at test time as an approximation to Bayesian inference. Multiple stochastic forward passes are performed with dropout layers active during test time. The softmax distributions $p(\mathbf{y|x, \hat{\omega}})$ obtained from these forward passes can then be used to compute either \emph{predictive entropy} (PE): $\mathbb{H}[\mathbb{E}_{\mathbf{\hat{\omega}}}[p(\mathbf{y|x, \hat{\omega}})]]$ or \emph{mutual information} (MI): $\mathbb{H}[\mathbb{E}_{\mathbf{\hat{\omega}}}[p(\mathbf{y|x, \hat{\omega}})]] - \mathbb{E}_{\mathbf{\hat{\omega}}}[\mathbb{H}[p(\mathbf{y|x, \hat{\omega}})]]$ \citep{houlsby2011bayesian} as measures of uncertainty. While MI is known to estimate epistemic uncertainty, PE captures both epistemic and aleatoric uncertainty \citep{Gal2016Uncertainty}. In our experiments, we implement MC dropout by activating the dropout layers in the DeepLab-v3+ architecture during test time. We don't insert new dropout layers. Finally, we use 5 stochastic forward passes for MC Dropout.
    
    \item \emph{\textbf{Deep Ensembles}} \citep{deep_ensembles} is a simple method where an ensemble of neural networks is trained. Similar to MC Dropout, both the PE as well as the MI from the ensemble predictions can be used to estimate uncertainty. In our experiments we use an ensemble of 3 DeepLab-v3+ models, all trained with identical architecture and training setup.
\end{enumerate}

\begin{figure*}[t]
    \centering
    \begin{subfigure}{0.30\linewidth}
        \includegraphics[width=\linewidth]{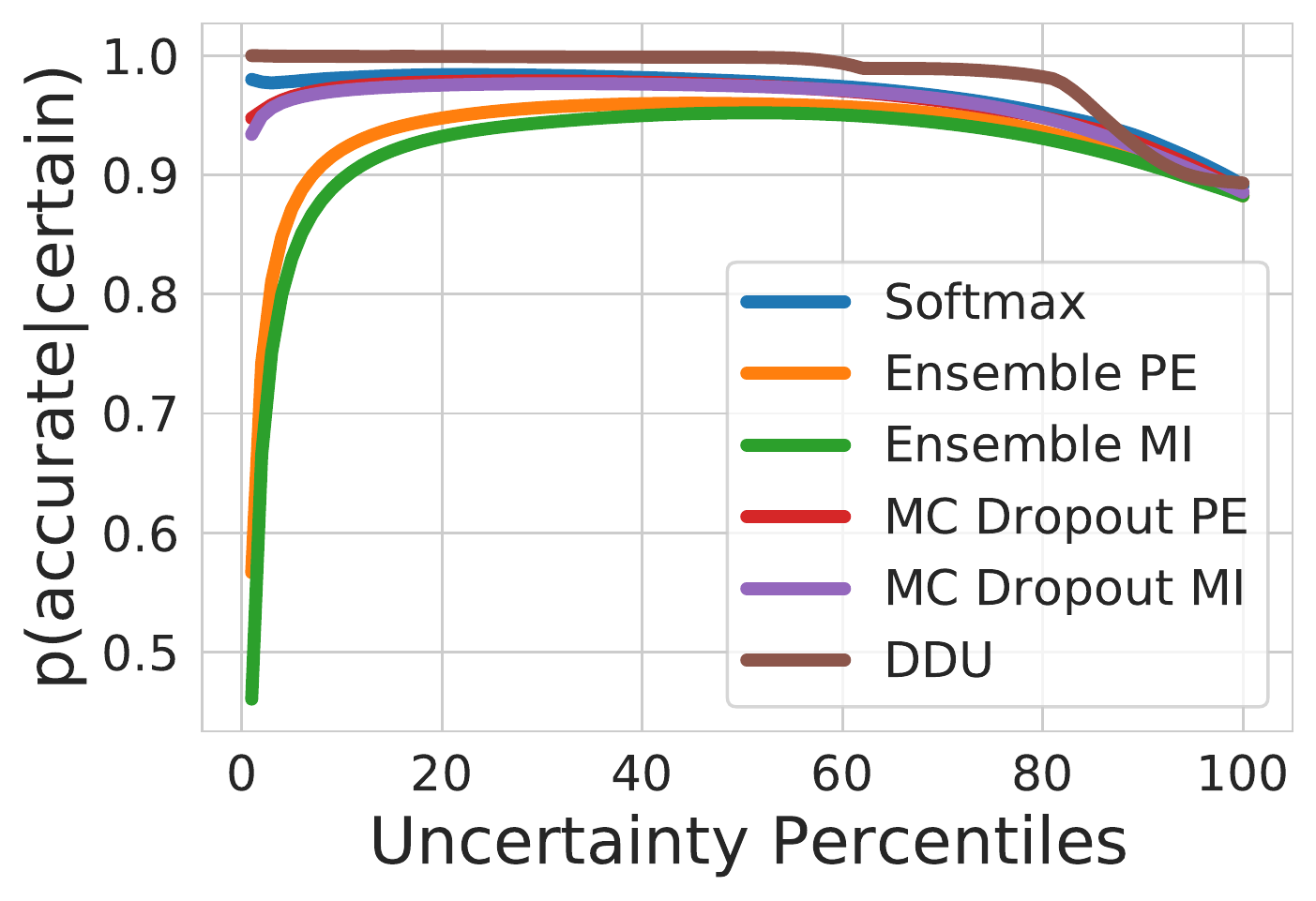}
        \label{fig:p_acc_cert}
    \end{subfigure}
    \begin{subfigure}{0.30\linewidth}
        \centering
        \includegraphics[width=\linewidth]{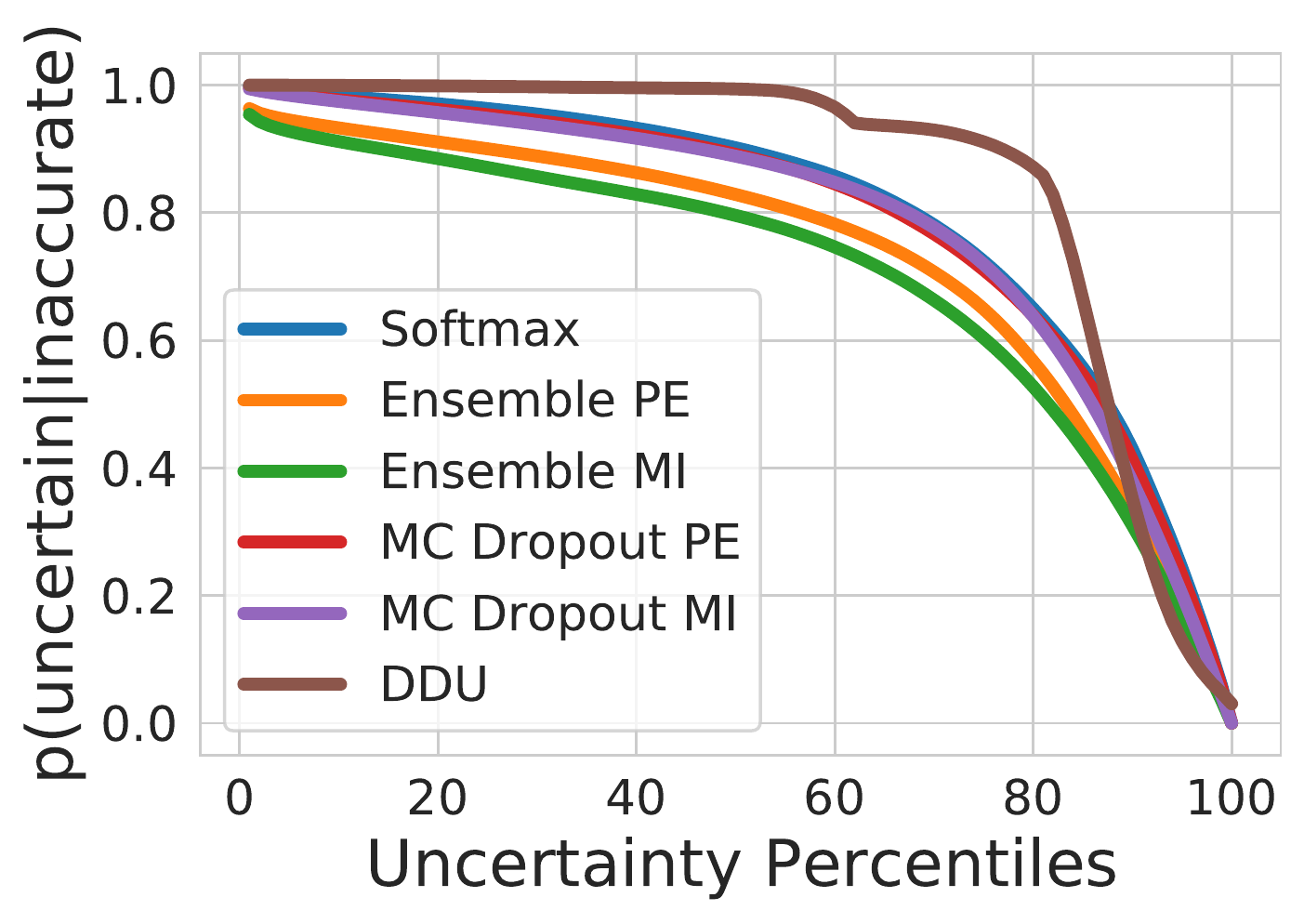}
        \label{fig:p_unc_inacc}
    \end{subfigure}
    \begin{subfigure}{0.30\linewidth}
        \centering
        \includegraphics[width=\linewidth]{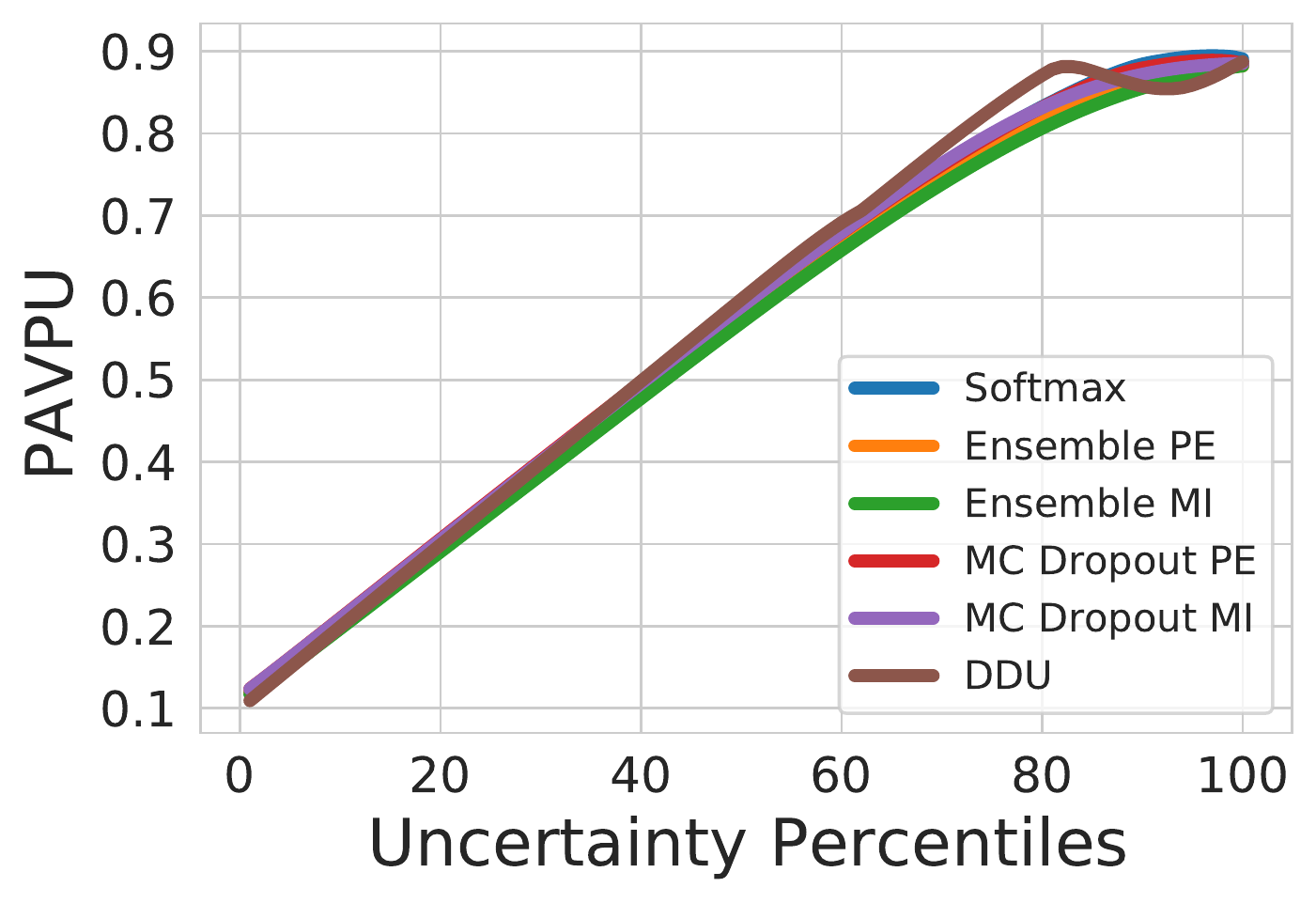}
        \label{fig:pavpu}
    \end{subfigure}
    \vspace{-3mm}
    \caption{
    Evaluation metrics: \emph{p(accurate|certain), p(uncertain|inaccurate)} and \emph{PAVPU} evaluated on different baselines on the PASCAL VOC validation set. DDU outperforms all other baselines.
    }
    \label{fig:3_metrics}
    \vspace{-3mm}
\end{figure*}

\textbf{Metrics for evaluation}: In order to evaluate the quality of uncertainty in semantic segmentation, we use the metrics proposed in \citep{mukhoti2018evaluating}: \emph{p(accurate|certain), p(uncertain|inaccurate)} and \emph{PAVPU}. The metric p(accurate|certain) measures the probability of a prediction being accurate given that the model is confident on the prediction. Similarly, p(uncertain|inaccurate) measures the probability of the model being uncertain on inaccurate predictions. PAVPU computes the probability of the model being confident on an accurate prediction or uncertain on an inaccurate one. A good model should ideally have high values on all these 3 metrics. Note that these metrics depend on a threshold for uncertainty, i.e., to define a prediction as certain or uncertain. Hence, they can be computed for different uncertainty thresholds. We plot the performance of all the baselines on these metrics for different uncertainty thresholds in \Cref{fig:3_metrics}.

\begin{figure*}[t!]
    \centering
    \begin{subfigure}{0.13\linewidth}
        \centering
        \includegraphics[width=\linewidth]{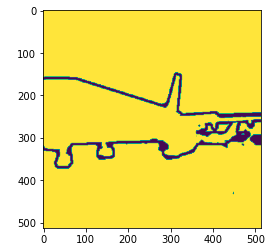}
    \end{subfigure}
    \begin{subfigure}{0.13\linewidth}
        \centering
        \includegraphics[width=\linewidth]{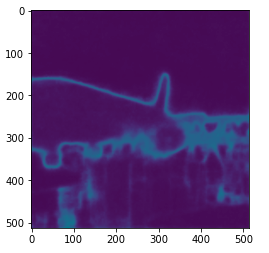}
    \end{subfigure}
    \begin{subfigure}{0.13\linewidth}
        \centering
        \includegraphics[width=\linewidth]{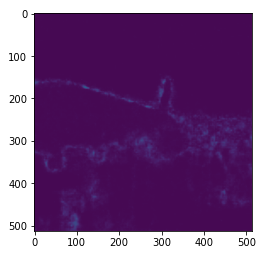}
    \end{subfigure}
    \begin{subfigure}{0.13\linewidth}
        \centering
        \includegraphics[width=\linewidth]{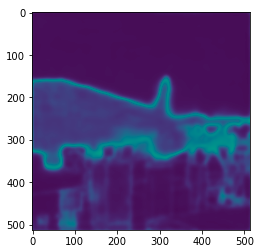}
    \end{subfigure}
    \begin{subfigure}{0.13\linewidth}
        \centering
        \includegraphics[width=\linewidth]{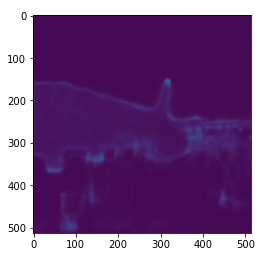}
    \end{subfigure}
    \begin{subfigure}{0.13\linewidth}
        \centering
        \includegraphics[width=\linewidth]{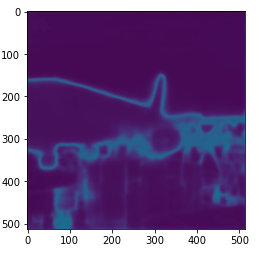}
    \end{subfigure}
    \begin{subfigure}{0.13\linewidth}
        \centering
        \includegraphics[width=\linewidth]{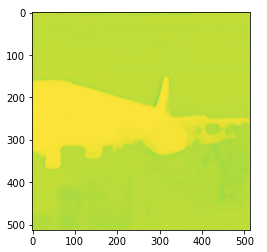}
    \end{subfigure}

    \begin{subfigure}{0.13\linewidth}
        \centering
        \includegraphics[width=\linewidth]{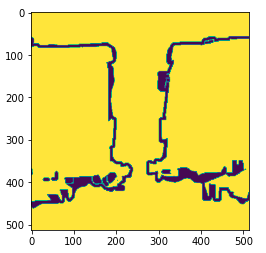}
    \end{subfigure}
    \begin{subfigure}{0.13\linewidth}
        \centering
        \includegraphics[width=\linewidth]{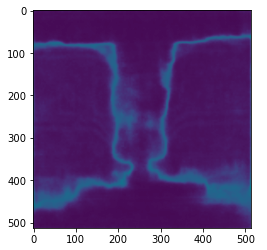}
    \end{subfigure}
    \begin{subfigure}{0.13\linewidth}
        \centering
        \includegraphics[width=\linewidth]{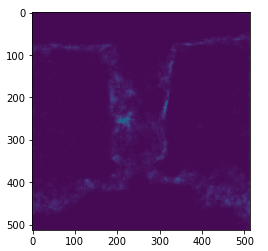}
    \end{subfigure}
    \begin{subfigure}{0.13\linewidth}
        \centering
        \includegraphics[width=\linewidth]{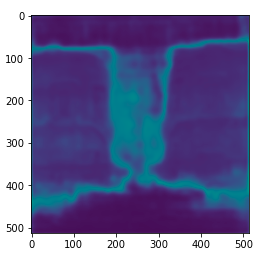}
    \end{subfigure}
    \begin{subfigure}{0.13\linewidth}
        \centering
        \includegraphics[width=\linewidth]{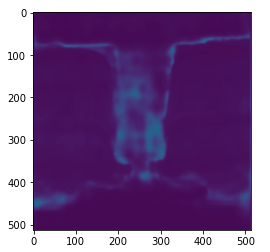}
    \end{subfigure}
    \begin{subfigure}{0.13\linewidth}
        \centering
        \includegraphics[width=\linewidth]{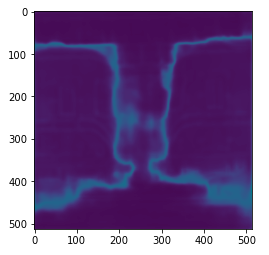}
    \end{subfigure}
    \begin{subfigure}{0.13\linewidth}
        \centering
        \includegraphics[width=\linewidth]{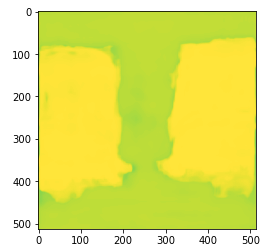}
    \end{subfigure}

    \begin{subfigure}{0.13\linewidth}
        \centering
        \includegraphics[width=\linewidth]{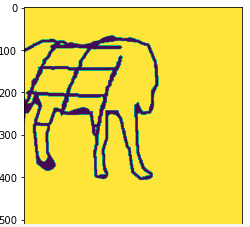}
    \end{subfigure}
    \begin{subfigure}{0.13\linewidth}
        \centering
        \includegraphics[width=\linewidth]{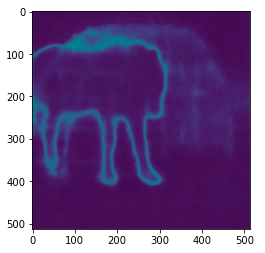}
    \end{subfigure}
    \begin{subfigure}{0.13\linewidth}
        \centering
        \includegraphics[width=\linewidth]{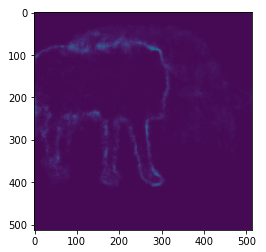}
    \end{subfigure}
    \begin{subfigure}{0.13\linewidth}
        \centering
        \includegraphics[width=\linewidth]{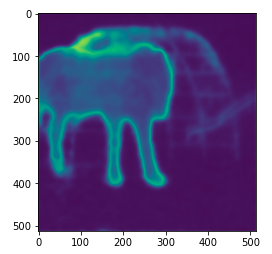}
    \end{subfigure}
    \begin{subfigure}{0.13\linewidth}
        \centering
        \includegraphics[width=\linewidth]{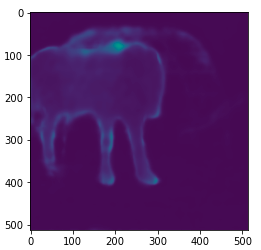}
    \end{subfigure}
    \begin{subfigure}{0.13\linewidth}
        \centering
        \includegraphics[width=\linewidth]{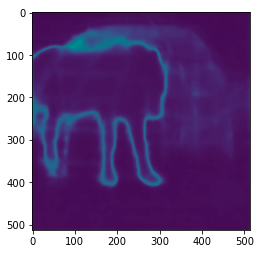}
    \end{subfigure}
    \begin{subfigure}{0.13\linewidth}
        \centering
        \includegraphics[width=\linewidth]{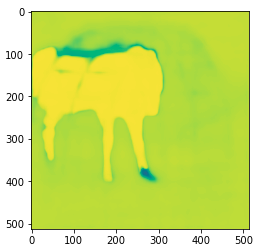}
    \end{subfigure}

    \begin{subfigure}{0.13\linewidth}
        \centering
        \includegraphics[width=\linewidth]{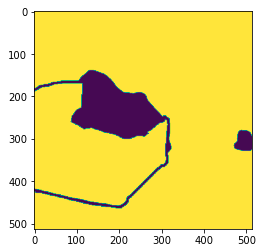}
        \caption{Accuracy}
    \end{subfigure}
    \begin{subfigure}{0.13\linewidth}
        \centering
        \includegraphics[width=\linewidth]{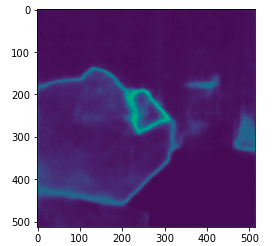}
        \caption{MCDO PE}
    \end{subfigure}
    \begin{subfigure}{0.13\linewidth}
        \centering
        \includegraphics[width=\linewidth]{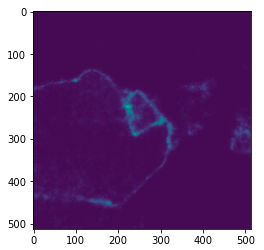}
        \caption{MCDO MI}
    \end{subfigure}
    \begin{subfigure}{0.13\linewidth}
        \centering
        \includegraphics[width=\linewidth]{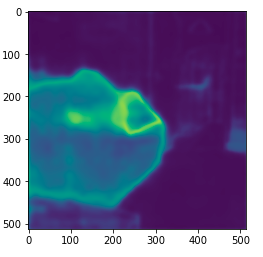}
        \caption{Ensemble PE}
    \end{subfigure}
    \begin{subfigure}{0.13\linewidth}
        \centering
        \includegraphics[width=\linewidth]{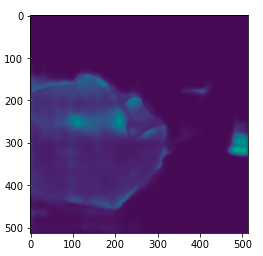}
        \caption{Ensemble MI}
    \end{subfigure}
    \begin{subfigure}{0.13\linewidth}
        \centering
        \includegraphics[width=\linewidth]{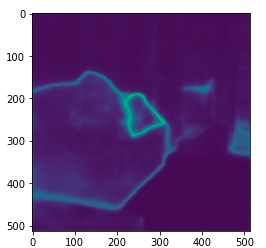}
        \caption{Entropy}
    \end{subfigure}
    \begin{subfigure}{0.13\linewidth}
        \centering
        \includegraphics[width=\linewidth]{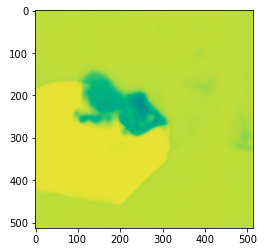}
        \caption{Density}
    \end{subfigure}
    \caption{
    \emph{Visualisation of different uncertainty baselines on samples from the PASCAL VOC validation set.} The first column captures pixel-wise accuracy with bright signifying accurate and dark, inaccurate. The second and third columns show predictive entropy (PE) and mutual information (MI) obtained from the MC Dropout (MCDO) baseline respectively, the fourth and fifth columns show the PE and MI from deep ensembles. \emph{The sixth column maps per-pixel softmax entropy which is the aleatoric uncertainty estimate of DDU, and finally the seventh column shows feature density, which is the epistemic component captured by DDU}. All the baselines save DDU density (last column on the right) capture uncertainty, i.e., the brighter, the more uncertain whereas DDU feature density captures confidence and hence brighter pixels signify more confident pixels and vice versa.
    }
    \label{fig:unc_vis}
\end{figure*}

In addition, we report the Pascal VOC 2012 validation set accuracy and the runtime of a single forward pass for all the baselines in \Cref{tab:runtime}. Note that a single forward pass for the MC Dropout baseline consists of five stochastic forward passes and a single forward pass from the ensemble involves getting predictions from three ensemble components. Finally, we visualise the uncertainty estimates from each baseline for four samples from the Pascal VOC 2012 val set in \Cref{fig:unc_vis}.

\textbf{Observations: } Firstly we note from \Cref{tab:runtime} that the runtime of DDU and a normal softmax model with values around $263$ and $275$ milliseconds respectively, are far lower than MC Dropout and Deep Ensembles. In fact, a single forward pass in MC Dropout requires around $1.57$ seconds on an Nvidia Quadro RTX 6000 GPU. Although these baselines have not been tuned for runtime, real-time latency requirements of around 200ms (i.e., 5 predictions a second) make adoption of time-consuming methods infeasible in real-life applications. Furthermore, note that the val set mIoU for all the models are very similar.

Secondly, from \Cref{fig:3_metrics}, we can see that DDU, having higher values on p(accurate|certain), p(uncertain|inaccurate) and PAVPU for most uncertainty thresholds, outperforms all other baselines on all these three metrics.

Finally, from \Cref{fig:unc_vis}, we note that DDU feature density captures epistemic uncertainty whereas softmax entropy captures aleatoric uncertainty. Note that aleatoric uncertainty is high on edges of objects as those are the regions with maximum ambiguity or observation noise. On the other hand, epistemic uncertainty is high on regions which are previously unseen (or less seen) by the model. In the first two samples (first two rows), the epistemic uncertainty is not high and aleatoric uncertainty is captured on the edges by softmax entropy. In the last sample (last row), the epistemic uncertainty is high for a big patch which is inaccurately predicted. DDU feature density for that entire patch is significantly lower whereas softmax entropy doesn't capture it and is only high on the edges.

\section{Conclusion}
\label{sec:conclusion}

In this paper, we show that Deep Deterministic Uncertainty (DDU) can be easily extended to the task of semantic segmentation. We find that fitting DDU to a semantic segmentation model with a fully convolutional architecture can be done in a pixel-independent fashion, thereby making its adoption relatively simple. Finally, with experiments on Pascal VOC 2012 using DeepLab-v3+, we observe that DDU outperforms other well-known methods of uncertainty quantification
without compromising on accuracy/mIoU and with the runtime of a single deterministic model.

\bibliography{example_paper}
\bibliographystyle{icml2021}

\newpage

\end{document}